\documentclass[runningheads]{llncs}
\usepackage{graphicx}
\usepackage{enumerate}
\usepackage[inline, shortlabels]{enumitem}
\usepackage{amsmath}
\usepackage{amssymb}
\usepackage{amstext} 
\usepackage{array}   
\newcolumntype{L}{>{$}l<{$}} 
\newcolumntype{C}{>{$}c<{$}} 
\usepackage{multirow}
\PassOptionsToPackage{hyphens}{url}\usepackage{hyperref}

\usepackage{subcaption}
\begin{document}
\title{Using 3D Convolutional Neural Networks to Learn Spatiotemporal Features for Automatic Surgical Gesture Recognition in Video
}
\titlerunning{3D CNNs for Automatic Surgical Gesture Recognition in Video}
%
\author{Isabel Funke\inst{1} \and Sebastian Bodenstedt\inst{1} \and Florian Oehme\inst{2} \and Felix von Bechtolsheim\inst{2} \and J\"urgen Weitz\inst{2, 3} \and Stefanie Speidel\inst{1, 3}}

\authorrunning{I. Funke et al.}
%

 \institute{Division  of  Translational  Surgical  Oncology,  National  Center  for  Tumor  Diseases~(NCT), Partner Site Dresden, Dresden, Germany \email{Firstname.Lastname@nct-dresden.de}\\ \and Department for Visceral, Thoracic and Vascular Surgery, University Hospital Carl Gustav Carus, TU Dresden, Dresden, Germany \and Centre for Tactile Internet with Human-in-the-Loop (CeTI), TU Dresden, Germany}
\maketitle              
\begin{abstract}
Automatically recognizing surgical gestures is a crucial step towards a thorough understanding of surgical skill. Possible areas of application include automatic skill assessment, intra-operative monitoring of critical surgical steps, and semi-automation of surgical tasks.
Solutions that rely only on the laparoscopic video and do not require additional sensor hardware are especially attractive as they can be implemented at low cost in many scenarios.
However, surgical gesture recognition based only on video is a challenging problem that requires effective means to extract both visual and temporal information from the video.
Previous approaches mainly rely on frame-wise feature extractors, either handcrafted or learned, which fail to capture the dynamics in surgical video.
To address this issue, we propose to use a 3D~Convolutional Neural Network (CNN) to learn spatiotemporal features from 
consecutive video frames.
We evaluate our approach on recordings of robot-assisted suturing on a bench-top model, which are taken from the publicly available JIGSAWS dataset.
Our approach achieves high frame-wise surgical gesture recognition accuracies of more than 84\%, outperforming comparable models that either extract only spatial features or model spatial and low-level temporal information separately.
For the first time, these results demonstrate the benefit of spatiotemporal CNNs for video-based surgical gesture recognition. 

\end{abstract}

\section{Introduction}
\label{sec:intro}

Surgical gestures~\cite{Ahmidi2017} are the basic elements of every surgical process.
Recognizing which surgical gesture is being performed is crucial for understanding the current surgical situation and for providing meaningful computer assistance to the surgeon.
Automatic surgical gesture recognition also offers new possibilities for surgical training. 
For example, it may enable a computer-assisted surgical training system  
to observe whether gestures are performed in the correct order or to identify with which gestures a trainee struggles the most.

Especially appealing is the exploitation of ubiquitous video feeds for surgical gesture recognition, such as the feed of the laparoscopic camera, which displays the surgical field in conventional and robot-assisted minimally invasive surgery.
The problem of \emph{video-based} surgical gesture recognition is formalized as follows: \\
A video of length $T$ is a sequence of video frames $v_t, t = 1, ..., T$.
The problem is to predict the gesture $g(t) \in \mathcal{G}$ performed at time $t$ for each $t = 1, ..., T$, where $\mathcal{G} = \lbrace{1, ..., G\rbrace}$ is the set of surgical gestures. 
Variations of surgical gesture recognition differ in the amount of information that is available to obtain an estimate $\hat{g}(t)$ of the current gesture, e.g.,
 \begin{enumerate*}[label=(\roman*)]
\item only the current video frame, i.e.,  $\hat{g}(t) = \hat{g}(v_t)$ (\emph{frame-wise} recognition), 
\item only frames up until the current timestep, i.e., $\hat{g}(t) = \hat{g}(v_k, ..., v_t), k \geq 1$ (\emph{on-line} recognition), or
\item the complete video, i.e., $\hat{g}(t) = \hat{g}(v_1, ..., v_T)$ (\emph{off-line} recognition).
\end{enumerate*}  

The main challenge in video-based surgical gesture recognition is the high dimensionality, high level of redundancy, and high complexity of video data.
State-of-the-art methods tackle the problem by
transforming video frames into feature representations, which are fed into temporal models that infer the sequence of gestures based on the input sequence.
These temporal models have been continuously improved in the last years, starting with variants of Hidden Markov Models~\cite{Tao2012} and Conditional Random Fields~\cite{lea2016st,tao2013surgical} and evolving into deep learning-based methods such as Recurrent Neural Networks~\cite{dipietro2016}, Temporal Convolutional Networks (TCN)~\cite{lea2016tcn}, and Deep Reinforcement Learning~(RL)~\cite{liu2018deep}. 

To obtain feature representations from video frames, early approaches compute bag-of-features histograms from feature descriptors extracted around space-time interest points or dense trajectories~\cite{tao2013surgical}.
More recently, \emph{Convolutional Neural Networks (CNNs})
became a popular tool for visual feature extraction.
For example, Lea et al. train a CNN (\emph{S-CNN}) for frame-wise gesture recognition~\cite{lea2016st} 
and use the latent video frame encodings as feature representations, which are further processed by a TCN for gesture recognition~\cite{lea2016tcn}.
A TCN combines 1D~convolutional filters with pooling and channel-wise normalization layers to hierarchically capture temporal relationships at low-, intermediate-, and high-level time scales.

Features extracted from individual video frames cannot represent the dynamics in surgical video, i.e., changes between adjacent frames. To alleviate this problem, Lea et al.~\cite{lea2016tcn} propose adding a number of difference images to the input fed to the S-CNN. For timestep $t$, difference images are calculated within a window of 2~seconds around frame $v_t$. 
Also, they suggest to use a \emph{spatiotemporal CNN (ST-CNN)}~\cite{lea2016st}, which applies a large temporal 1D~convolutional filter to the latent activations obtained by a S-CNN. 
In contrast, we propose to use a 3D~CNN to learn spatiotemporal features from stacks of consecutive video frames,
thus modeling the temporal evolution of video frames directly.

To the best of our knowledge, we are the first to design a 3D~CNN for surgical gesture recognition that predicts gesture labels for consecutive frames of surgical video. 
An evaluation on the suturing task of the publicly available JIGSAWS~\cite{Ahmidi2017} dataset demonstrates the superiority of our approach compared to 2D~CNNs that estimate surgical gestures based on spatial features extracted from individual video frames. 
Averaging the dense predictions of the 3D~CNN over time even
achieves compelling frame-wise gesture recognition accuracies of over 84\%. 
Source code can be accessed at \url{https://gitlab.com/nct_tso_public/surgical_gesture_recognition}.

\section{Methods}

In the following, we detail the architecture and training procedure of the proposed 3D~CNN for video-based surgical gesture recognition.

\subsection{Network Architecture}
Ji et al.~\cite{Ji2013} proposed \emph{3D~CNNs} as a natural extension of well-known (2D)~CNNs. While 2D~CNNs apply 2D~convolutions and 2D~pooling kernels to extract features along the spatial dimensions of a video frame~$v \in \mathbb{R}^{C \times H \times W}$, 3D~CNNs apply 3D~convolutions and 3D~pooling kernels to extract features along the spatial and temporal dimensions of a stack of video frames~$\vartheta = [v_k, v_{k + 1}, ..., v_{k + L - 1}] \in \mathbb{R}^{C \times L \times H \times W}$.
Recently, Carreira et al.~\cite{Carreira2017} suggested to create 3D~CNN architectures by \emph{inflating} etablished deep 2D~CNN architectures
along the temporal dimension. This basically means that all $N \times N$~kernels are expanded into their cubic $N \times N \times N$ counterparts. 

The proposed 3D~CNN for surgical gesture recognition is based on 3D~ResNet-18~\cite{hara2017learning}, which is created by inflating an 18-layer residual network~\cite{He2016}.
Input to the network are stacks of 16 consecutive video frames (as proposed in~\cite{hara2017learning}) with a resolution of $224 \times 224$ pixels.
More precisely, to obtain an estimate $\hat{g}(t)$ of the gesture being performed at time $t$, we feed the video snippet $\vartheta_t = (v_{t - 15}, ..., v_{t - 1}, v_t)$ to the network.
Because we process the video at 5~fps, the network can refer to the previous three seconds of video in order to infer $\hat{g}(t)$.
At this point, we abstain from feeding future video frames to the network so that the method is applicable for online gesture recognition.

The original 3D~ResNet-18 architecture is designed to predict one distinct action label per video snippet using a one-hot encoding. In contrast, surgical gesture recognition is a dense labeling problem, where each frame $v_k$ of a video snippet has a distinct label $g(k)$. This means that one video snippet may contain frames that belong to different gestures.
To account for this,
we adapt our network to output dense gesture label estimates $\hat{\gamma}_t = (\hat{g}_t(t - 15), ..., \hat{g}_t(t - 1), \hat{g}_t(t)) \in \mathbb{R}^{G \times 16}$. Here, $G$ denotes the  number of distinct surgical gestures. The component $\hat{g}_t(t - i), i = 0, ..., 15,$ of $\hat{\gamma}_t$ is the estimate for gesture label $g(t - i)$, obtained at time~$t$.

Specifically, we adapt the max pooling layer of 3D~ResNet-18 so that downsampling is only performed along the spatial dimensions. 
Thus, the feature maps after the final average pooling layer have a dimension of $512 \times 2$. This is upsampled to the output dimension $G \times 16$ using a transposed 1D~convolution ($\text{conv}^T$) with kernel size 11 and stride 5.

An overview of the network architecture is given in table~\ref{tab:architecture}.  
The input is downsampled in the initial convolutional and max pooling layers and then passed through a number of residual blocks. When convolutions are applied with stride~2 to downsample feature maps, the number of feature maps is doubled. For details on residual blocks, please see the original papers~\cite{hara2017learning,He2016}.
We apply batch normalization and the ReLU non-linearity after each convolutional layer. An exception is the final transposed convolution, which is normalized using a softmax layer.

\begin{table}[t]
\caption{Network architecture. For each layer type, we specify kernel size, number of output feature maps, and, if applicable, stride or number of residual blocks. Square brackets indicate shortcut connections.}
\label{tab:architecture}      
\centering
\setlength{\tabcolsep}{8pt}
\begin{tabular}{L L L}
\hline\noalign{\medskip}
\text{layer type} & & \text{output size} \\
\noalign{\smallskip}\hline\noalign{\smallskip}
\text{conv} & ~7 \times 7 \times 7, 64, ~\text{stride} (1, 2, 2) & 16 \times 112 \times 112\\
\noalign{\smallskip}\hline\noalign{\smallskip}
\text{max pool} & ~1 \times 3 \times 3, 64, ~\text{stride} (1, 2, 2) & 16 \times 56 \times 56 \\
\noalign{\smallskip}
\text{res block} &
	\begin{bmatrix}
       ~3 \times 3 \times 3, ~64~\\[0.3em]
       ~3 \times 3 \times 3, ~64~\\[0.3em]
     \end{bmatrix}
     \times 2 & \\
\noalign{\smallskip}\hline\noalign{\smallskip}
\text{res block} &
	\begin{bmatrix}
       ~3 \times 3 \times 3, 128~\\[0.3em]
       ~3 \times 3 \times 3, 128~\\[0.3em]
     \end{bmatrix}
     \times 2
& 8 \times 28 \times 28 \\
\noalign{\smallskip}\hline\noalign{\smallskip}		
\text{res block} &	
	\begin{bmatrix}
       ~3 \times 3 \times 3, 256~\\[0.3em]
       ~3 \times 3 \times 3, 256~\\[0.3em]
     \end{bmatrix}
     \times 2
& 4 \times 14 \times 14 \\
\noalign{\smallskip}\hline\noalign{\smallskip}	
\text{res block} &
	\begin{bmatrix}
       ~3 \times 3 \times 3, 512~\\[0.3em]
       ~3 \times 3 \times 3, 512~\\[0.3em]
     \end{bmatrix}
     \times 2
& 2 \times 7 \times 7 	\\	
\noalign{\smallskip}\hline\noalign{\smallskip}	
\text{avg pool} & ~1 \times 7 \times 7, 512 & 2 \\
\noalign{\smallskip}\hline\noalign{\smallskip}	
\text{conv}^{T} & ~11, G ~\text{stride 5} & 16 \\
\noalign{\smallskip}\hline\noalign{\smallskip}	
\end{tabular}
\end{table}

\subsection{Network Training}
\label{sec:training}

We train our 3D~CNN on video snippets $\vartheta_t = (v_{t - 15}, ..., v_{t - 1}, v_t)$ to predict the corresponding ground truth gesture labels $\gamma_t = (g(t - 15), ..., g(t - 1), g(t))$.
Therefore, we minimize the loss $\mathcal{L}(\gamma_t, \hat{\gamma}_t) = \sum_{i=0}^{15}\omega_i \mathcal{L}_{CE}(g(t-i), \hat{g}_t(t-i))$, where $\mathcal{L}_{CE}$ denotes the cross entropy loss.
We found it to be beneficial to penalize the errors made on more current predictions harder and therefore train with weighting factors $\omega_i = \frac{(16 - i)^2}{\sum_{i=0}^{15} (16 - i)^2}$.

Because of their large number of parameters, 3D~CNNs are difficult to train, especially on small datasets~\cite{hara2017learning}.
Thus, it is important to begin training from a suitable initialization of network parameters. We investigate two approaches for network initialization:
\begin{enumerate*} [label=(\roman*)]
	\item Initializing the network with parameters obtained by training on Kinetics~\cite{Carreira2017}, one of the largest human action datasets available so far. For this, a publicly available pretrained 3D~ResNet-18 model\footnote{\url{https://github.com/kenshohara/3D-ResNets-PyTorch\#pre-trained-models}}~\cite{hara2017learning} is used.
	\item Bootstrapping network parameters from an ImageNet-pretrained 2D~ResNet-18 model that was further trained on individual video frames to perform frame-wise gesture recognition.
	As described in~\cite{Carreira2017}, the 3D~filters of the 3D~ResNet-18 are initialized by repeating the weights of the corresponding 2D~filters $N$~times along the temporal dimension and then dividing them by~$N$.
\end{enumerate*}

During training, we sample video snippets $\vartheta_t$ at random temporal positions~$t$ from the training videos.
Per epoch, we sample about 3000~snippets in a class-balanced manner, which means that we ensure that each gesture $g \in \mathcal{G}$ is represented equally in the set of sampled snippets.
For data augmentation, we use scale jittering and corner cropping as proposed in~\cite{Wang2016}. 
Here, all frames within one training snippet are augmented in the same manner.
We train the 3D~CNN for 250 epochs using the Adam~\cite{Kingma2015} optimizer with a batch size of 32 and an initial learning rate of $2.5 \cdot 10^{-4}$. The learning rate is divided by factor~5 every 50~epochs. 
Our 3D~CNN implementation is based on code\footnote{\url{https://github.com/kenshohara/3D-ResNets-PyTorch}} provided by~\cite{hara2017learning}.

\section{Evaluation}

We evaluate our approach on 39 videos of robot-assisted suturing tasks performed on a bench-top model, which are taken from the \emph{JHU-ISI Gesture and Skill Assessment Working Set (JIGSAWS)}~\cite{Ahmidi2017}.
The recorded tasks were performed by eight participants with varying surgical experience. 
The videos were annotated with surgical gestures such as \emph{positioning the tip of the needle} or \emph{pushing needle through the tissue}. In total, $G = 10$ different gestures are used.
We follow the leave-one-user-out (LOUO) setup for cross-validation as defined in~\cite{Ahmidi2017}. Thus, for each experiment, we train one model per left-out user.

We report the following evaluation metrics: 
\begin{enumerate*}[label=(\roman*)]
	\item Frame-wise accuracy, i.e., the ratio of correctly predicted gesture labels in a video.
	\item Average $F_1$ score, where we calculate the $F_1$ score, i.e., the harmonic mean of precision and recall, with respect to each gesture class and average the results over all classes. 
	\item Edit score, as proposed in~\cite{lea2016st}, which employs the Levenshtein distance to assess the quality of predicted gesture segments.
	\item Segmental $F_1$ score with threshold~10\% ($F_1@10$), as proposed in~\cite{lea2017temporal}. Here, a predicted gesture segment is considered a true positive if its intersection with the corresponding ground truth segment is over 10\%, and the $F_1$ score is calculated regarding the total number of true positives, false positives, and false negatives. 	
\end{enumerate*}  
For each experiment, evaluation metrics are calculated for every video in the dataset and then averaged.

\begin{table}[t]
\caption{Experimental results on the suturing task of JIGSAWS.
The column captioned with \emph{look ahead} indicates how much future video a method needs to see to estimate the current gesture. 
Evaluation measures that were not reported in related work are denoted as --. 
All measures are given in \%.}
\label{tab:results}   
\setlength{\tabcolsep}{8pt}
\centering   
\begin{tabular}{llllll}
\hline\noalign{\medskip}
\multirow{2}{*}{Method}		& Look  & \multirow{2}{*}{Acc} & \multirow{2}{*}{Avg. $F_1$} & \multirow{2}{*}{Edit} & \multirow{2}{*}{$F_1@{10}$} \\
							& ahead &  & & & \\
\noalign{\smallskip}\hline\noalign{\smallskip}
\textit{Evaluation at 5 fps} & & & & & \\
\hspace{5pt} 2D ResNet-18 & 0s & 79.9 & 73.3 & 41.4 & 55.4 \\
\hspace{5pt} 3D CNN (B)	& 0s & 79.9 & 73.7 & 64.0 & 75.2 \\
\hspace{5pt} 3D CNN (K)	& 0s & 81.8 & 75.8 & 58.7 & 71.1 \\
\hspace{5pt} 3D CNN (B) + window & 3s & 84.0 & \textbf{78.4} & \textbf{80.7} & \textbf{87.2} \\
\hspace{5pt} 3D CNN (K) + window & 3s & \textbf{84.2} & \textbf{78.4} & 80.0 & 87.1 \\
\noalign{\smallskip}\hline\noalign{\smallskip}
\textit{Evaluation at 10 fps} & & & & & \\
\hspace{5pt} S-CNN~\cite{lea2016tcn}	& 1s 		& 74.0 & -- & 37.7 & -- \\
\hspace{5pt} S-CNN + TCN, $C = 10$, causal & 1s	& 76.8 & 71.5 & 57.3 & 69.6 \\
\hspace{5pt} S-CNN + TCN, $C = 10$ & 3s	& 76.1 & 69.9 & 68.2 & 77.9 \\
\hspace{5pt} ST-CNN~\cite{lea2016tcn}	& 10s		& 77.7 & -- & 68.0 & -- \\
\hspace{5pt} S-CNN + TCN, $C = 75$ & 22.5s	& 81.4 & 77.6 & 84.9 & 89.6 \\
\hspace{5pt} S-CNN + TCN + Deep RL~\cite{liu2018deep} & -- & 81.4 & -- & \textbf{88.0} & \textbf{92.0} \\
\hspace{5pt} 2D ResNet-18	& 0s & 79.5 & 73.1 & 30.6 & 44.2 \\
\hspace{5pt} 3D CNN (B)	& 0s & 79.5 & 73.6 & 49.5 & 62.8 \\
\hspace{5pt} 3D CNN (K)	& 0s & 81.3 & 75.1 & 46.3 & 60.1 \\
\hspace{5pt} 3D CNN (B) + window & 3s & 84.0 & 78.6 & 80.6 & 87.0 \\
\hspace{5pt} 3D CNN (K) + window & 3s & \textbf{84.3} & \textbf{78.6} & 80.0 & 87.0 \\
\noalign{\smallskip}\hline
\bigskip
\end{tabular}
\end{table}

As baseline experiment, we train a 2D~ResNet-18~\cite{He2016}, i.e., the 2D counterpart to the proposed 3D~CNN, for frame-wise gesture recognition. Here, we follow the training procedure described in section~\ref{sec:training} except for the fact that we train on video snippets of size 1, i.e., individual video frames. The 2D~ResNet-18 is initialized with ImageNet-pretrained weights.
 
Additionally, we perform two experiments where we train the proposed 3D CNN for surgical gesture recognition: one where we initialize the 3D~CNN with Kinetics-pretrained weights \emph{(3D~CNN (K))} and one where we bootstrap weights from a pretrained 2D~ResNet-18 as described in section~\ref{sec:training} \emph{(3D~CNN (B))}.
To account for the stochastic nature of CNN optimization, we repeat the three experiments four times and report the averaged results. 
For the \emph{3D~CNN~(B)} experiment, we initialize the models in the $i^{\text{th}}$ experiment repetition by bootstrapping weights from the corresponding 2D~ResNet-18 models (with respect to the LOUO splits) that were trained during the $i^{\text{th}}$ repetition of the baseline experiment.

We evaluate the trained 3D~CNN models either snippet-wise or in combination with a sliding window (\emph{+ window}).
For snippet-wise evaluation, the estimated gesture label $\hat{g}(t)$ at time $t$ is simply $\hat{g}_t(t)$.
With the sliding window approach, we accumulate the dense predictions of the 3D~CNN  over time.
This yields the overall estimate $\bar{\hat{g}}(t) = \sum_{i = 0}^{15} \hat{g}_{t + i}(t)$ for the gesture at time $t$.
To obtain $\bar{\hat{g}}(t)$, information of 15 future time steps is used, which corresponds to the next three seconds of video. 

To make comparisons to prior studies possible, we additionally evaluate the 2D~ResNet-18 and the {3D~CNN models at 10 fps. This means that we extract video snippets at 10~Hz, instead of 5~Hz, from the video. For the 3D~CNNs, the individual snippets still consist of 16 frames sampled at 5~fps. To apply the sliding window approach, we temporally upsample the prediction $\hat{\gamma}_t \in \mathbb{R}^{G \times 16}$ to $\tilde{\gamma}_t \in \mathbb{R}^{G \times 32}$, where 
$
\tilde{\gamma}_t[j] =\begin{cases}
  \hat{\gamma}_t[0],  & \text{if }j=0,\\
 0.5 \cdot \hat{\gamma}_t[\left \lfloor{\frac{j - 1}{2}}\right \rfloor] + 0.5 \cdot \hat{\gamma}_t[\left \lceil{\frac{j - 1}{2}}\right \rceil], & \text{if }j = 1, ..., 31.
\end{cases}
$

The experimental results are listed in table~\ref{tab:results}. For comparison, we state the results of some previous methods that were described in section~\ref{sec:intro}. Further experiments can be found in the supplementary document.

\emph{S-CNN + TCN} refers to the method where spatial features are extracted from video frames using a S-CNN and fed to a TCN that predicts surgical gestures~\cite{lea2017temporal,lea2016tcn}.
Here, the results were reproduced using the ED-TCN architecture described in~\cite{lea2017temporal} with 2~layers and temporal filter size $C$. For causal evaluation, filters are applied from $v_{t - C}$ to $v_t$ instead of $v_{t - C/2}$ to $v_{t + C/2}$. 
We use source code\footnote{\url{https://github.com/colincsl/TemporalConvolutionalNetworks}} provided by the authors of~\cite{lea2017temporal,lea2016tcn}. 
The reported results are averaged over four LOUO cross-validation runs.

\section{Discussion}
As can be seen in table~\ref{tab:results}, 
the proposed variant of 3D~ResNet-18 for snippet-wise gesture recognition yields comparable or better frame-wise evaluation results (accuracy and average $F_1$) and considerably better segment-based evaluation results (edit score and $F_1@10$) compared to the 2D~counterpart.
This demonstrates the benefit of modeling several consecutive video frames to capture the temporal evolution of video.

Accumulating the 3D~CNN predictions using a sliding window with a duration of three seconds provides a further boost to recognition performance. 
Not only does the sliding window approach produce better gesture segments, it also improves frame-wise accuracies.
Considering future video snippets most likely helps to resolve ambiguities in individual snippets.

Minor differences can be observed between both network initialization variants, Kinetics pretraining (K) and 2D~weight bootstrapping (B): while pretraining on Kinetics yields higher frame-wise accuracies, 
the other approach yields better gesture segments. In combination with the sliding window the differences are marginal.

When testing at 10~fps instead of 5~fps, we observe a notable degradation of the segment-based measures for both the 2D ResNet-18 and the 3D variants. Most likely, the high evaluation frequency enhances noise in the gesture predictions, which is penalized by the edit score and the $F_1@{10}$ metric.  
For the 3D~CNNs, this effect can be alleviated by filtering with the sliding window.

Compared to the ST-CNN, the 3D~CNN yields considerably better results with regards to all evaluation metrics when being evaluated with the sliding window approach.
Apparently, for the given task, modeling spatiotemporal features in video snippets achieves better results than modeling spatial and temporal information separately, as is the case for the ST-CNN.

In combination with the sliding window, the proposed 3D~CNN also outperforms the state-of-the-art methods \emph{S-CNN + TCN} and \emph{S-CNN + TCN + Deep RL} in terms of accuracy and average $F_1$. 
These methods apply very long temporal filters
while the proposed approach only processes a few seconds of video to estimate the current gesture. Thus,
it is surprising that the quality of gesture segments, as measured by edit score and $F_1@10$, is almost equal.

Note that the proposed method operates with a delay of only 3 seconds and 
can therefore provide information, such as feedback in a surgical training scenario, in a more timely manner than methods with a longer \emph{look ahead} time.

\section{Conclusion}
We present a 3D~CNN to predict dense gesture labels for surgical video.
The conducted experiments demonstrate the benefits of using an inherently spatiotemporal model to extract features from consecutive video frames. 
Future work will investigate options for combining spatiotemporal feature extractors with models that capture high-level temporal dependencies, such as LSTMs or TCNs. 

\subsubsection{Acknowledgements}
The authors thank Colin Lea for sharing code and precomputed S-CNN features to reproduce results from~\cite{lea2016tcn} as well as the Helmholtz-Zentrum Dresden-Rossendorf~(HZDR) for granting access to their GPU cluster.

%


\bibliographystyle{splncs04}
\bibliography{mybibliography}

\begin{thebibliography}{10}
\providecommand{\url}[1]{\texttt{#1}}
\providecommand{\urlprefix}{URL }
\providecommand{\doi}[1]{https://doi.org/#1}

\bibitem{Ahmidi2017}
Ahmidi, N., Tao, L., Sefati, S., Gao, Y., Lea, C., Haro, B.B., et~al.: A
  dataset and benchmarks for segmentation and recognition of gestures in
  robotic surgery. IEEE Trans Biomed Eng  \textbf{64}(9),  2025--2041 (2017)

\bibitem{Carreira2017}
Carreira, J., Zisserman, A.: Quo vadis, action recognition? {A} new model and
  the {Kinetics} dataset. In: CVPR. pp. 4724--4733. IEEE (2017)

\bibitem{dipietro2016}
DiPietro, R., Lea, C., Malpani, A., Ahmidi, N., Vedula, S.S., Lee, G.I.,
  et~al.: Recognizing surgical activities with recurrent neural networks. In:
  MICCAI. pp. 551--558. Springer, Cham (2016)

\bibitem{hara2017learning}
Hara, K., Kataoka, H., Satoh, Y.: Learning spatio-temporal features with {3D}
  residual networks for action recognition. In: ICCV-W. pp. 3154--3160. IEEE
  (2017)

\bibitem{He2016}
He, K., Zhang, X., Ren, S., Sun, J.: Deep residual learning for image
  recognition. In: CVPR. pp. 770--778. IEEE (2016)

\bibitem{Ji2013}
Ji, S., Xu, W., Yang, M., Yu, K.: {3D} convolutional neural networks for human
  action recognition. IEEE Trans Pattern Anal Mach Intell  \textbf{35}(1),
  221--231 (2013)

\bibitem{Kingma2015}
Kingma, D.P., Ba, J.: Adam: {A} method for stochastic optimization. In: ICLR
  (2015)

\bibitem{lea2017temporal}
Lea, C., Flynn, M.D., Vidal, R., Reiter, A., Hager, G.D.: Temporal
  convolutional networks for action segmentation and detection. In: CVPR. pp.
  156--165. IEEE (2017)

\bibitem{lea2016st}
Lea, C., Reiter, A., Vidal, R., Hager, G.D.: Segmental spatiotemporal {CNN}s
  for fine-grained action segmentation. In: ECCV. pp. 36--52. Springer, Cham
  (2016)

\bibitem{lea2016tcn}
Lea, C., Vidal, R., Reiter, A., Hager, G.D.: Temporal convolutional networks: A
  unified approach to action segmentation. In: ECCV-W. pp. 47--54. Springer,
  Cham (2016)

\bibitem{liu2018deep}
Liu, D., Jiang, T.: Deep reinforcement learning for surgical gesture
  segmentation and classification. In: MICCAI. pp. 247--255. Springer, Cham
  (2018)

\bibitem{Tao2012}
Tao, L., Elhamifar, E., Khudanpur, S., Hager, G.D., Vidal, R.: Sparse hidden
  markov models for surgical gesture classification and skill evaluation. In:
  IPCAI. pp. 167--177. Springer, Berlin, Heidelberg (2012)

\bibitem{tao2013surgical}
Tao, L., Zappella, L., Hager, G.D., Vidal, R.: Surgical gesture segmentation
  and recognition. In: MICCAI. pp. 339--346. Springer, Berlin, Heidelberg
  (2013)

\bibitem{Wang2016}
Wang, L., Xiong, Y., Wang, Z., Qiao, Y., Lin, D., Tang, X., et~al.: Temporal
  segment networks: Towards good practices for deep action recognition. In:
  ECCV. pp. 20--36. Springer, Cham (2016)

\end{thebibliography}

\title{Supplementary}
\subtitle{Using 3D Convolutional Neural Networks to Learn Spatiotemporal Features for Automatic Surgical Gesture Recognition in Video}
\titlerunning{Supplementary -- 3D CNNs for Automatic Surgical Gesture Recognition}
%
\author{Isabel Funke\inst{1}\and Sebastian Bodenstedt\inst{1} \and Florian Oehme\inst{2} \and Felix von Bechtolsheim\inst{2} \and J\"urgen Weitz\inst{2,3} \and Stefanie Speidel\inst{1,3}}

\authorrunning{I. Funke et al.}
%

 \institute{Division  of  Translational  Surgical  Oncology,  National  Center  for  Tumor  Diseases~(NCT), Partner Site Dresden, Dresden, Germany\\ \and Department for Visceral, Thoracic and Vascular Surgery, University Hospital Carl Gustav Carus, TU Dresden, Dresden, Germany \and Centre for Tactile Internet with Human-in-the-Loop (CeTI), TU Dresden, Germany}
\maketitle              

\begin{figure}[h]
\centering
\includegraphics[width=0.62\textwidth]{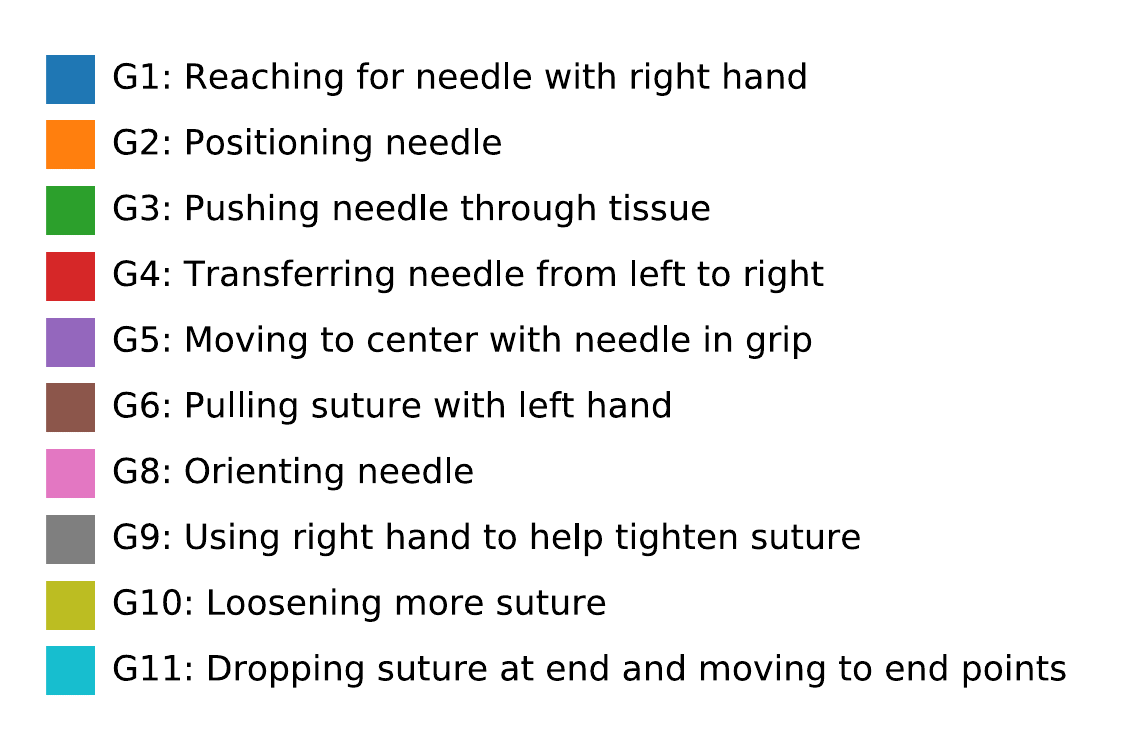}
\caption{Surgical gestures defined for the JIGSAWS suturing task} 
\label{gestures}
\end{figure}

\begin{table}[h]
\caption{Inference times. We report the time required to process a batch of size~1 using either the 2D or the 3D~CNN. The results are averaged over 2500 measurements each, taken on a Nvidia GeForce GTX 1080.}
\label{tab:inference_times}   
\setlength{\tabcolsep}{8pt}
\centering   
\begin{tabular}{lll}
\hline\noalign{\medskip}
		& 2D ResNet-18  & 3D CNN \\
\noalign{\smallskip}\hline\noalign{\smallskip}\noalign{\smallskip} 
Input size & 3 $\times$ 224 $\times$ 224 & 3 $\times$ 16 $\times$ 224 $\times$ 224\\
Output size & 10 & 10 $\times$ 16 \\
Inference time (ms) & 3.8 $\pm$ 0.05 & 25.5 $\pm$ 0.12 \\
\noalign{\smallskip}\hline
\bigskip
\end{tabular}
\end{table}

\begin{figure}[h]
\label{qualitativeResults}
\centering
\begin{subfigure}[b]{\textwidth}
   \includegraphics[width=1\linewidth]{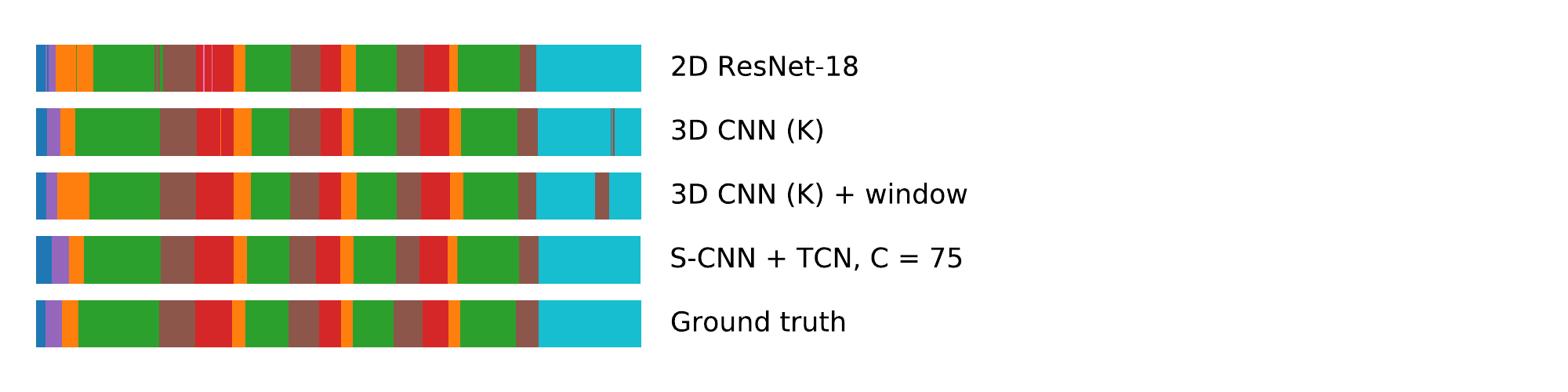}
   \label{fig:Ng1} 
\end{subfigure}

\begin{subfigure}[b]{\textwidth}
   \includegraphics[width=1\linewidth]{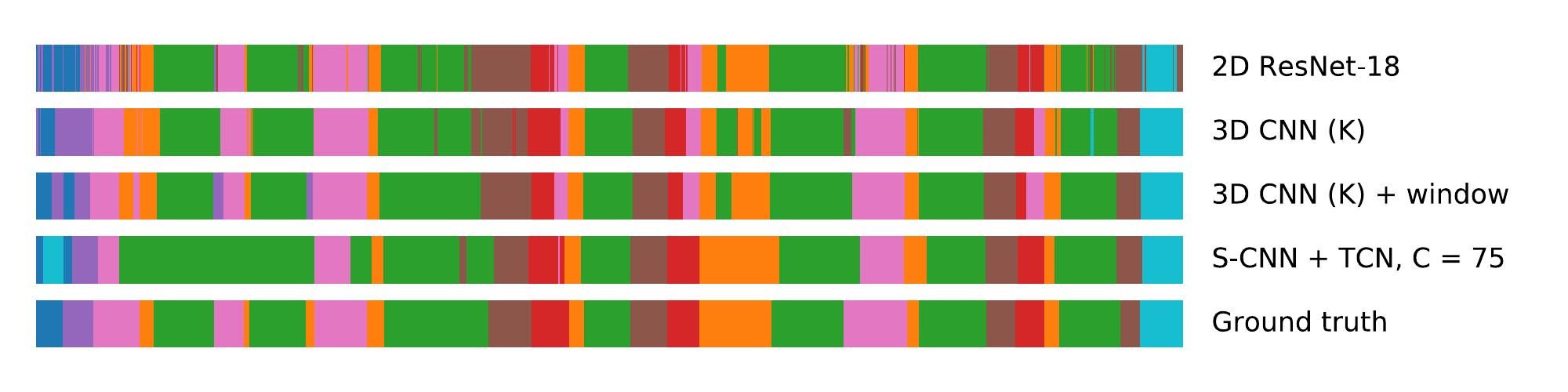}
   \label{fig:Ng2}
\end{subfigure}

\begin{subfigure}[b]{\textwidth}
   \includegraphics[width=1\linewidth]{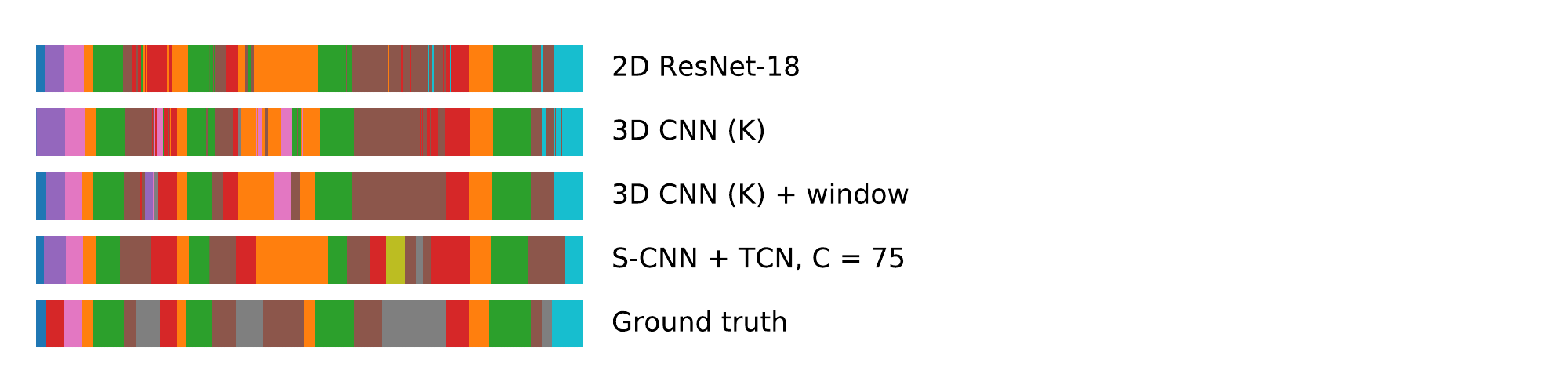}
   \label{fig:Ng2}
\end{subfigure}

\caption{Qualitative results. We depict the surgical gesture estimates obtained by our method (3D~CNN (K), 3D~CNN (K) + window), which uses a 3D~CNN to extract spatiotemporal features from video. For comparison, we illustrate the predictions obtained using a 2D~ResNet-18 or a 2D~CNN in combination with a Temporal Convolutional Network (S-CNN + TCN, C = 75). Evaluation was conducted at 10 fps. The presented sequences are those where the proposed method (3D~CNN~(K)) achieves highest (top), medium (middle), and lowest (bottom) recognition accuracies. Colors encode gesture labels, see Fig.~\ref{gestures}.}
\end{figure}

\begin{figure}[h]
\centering
\includegraphics[width=1.0\textwidth, trim={0 4cm 10.7cm 0},clip]{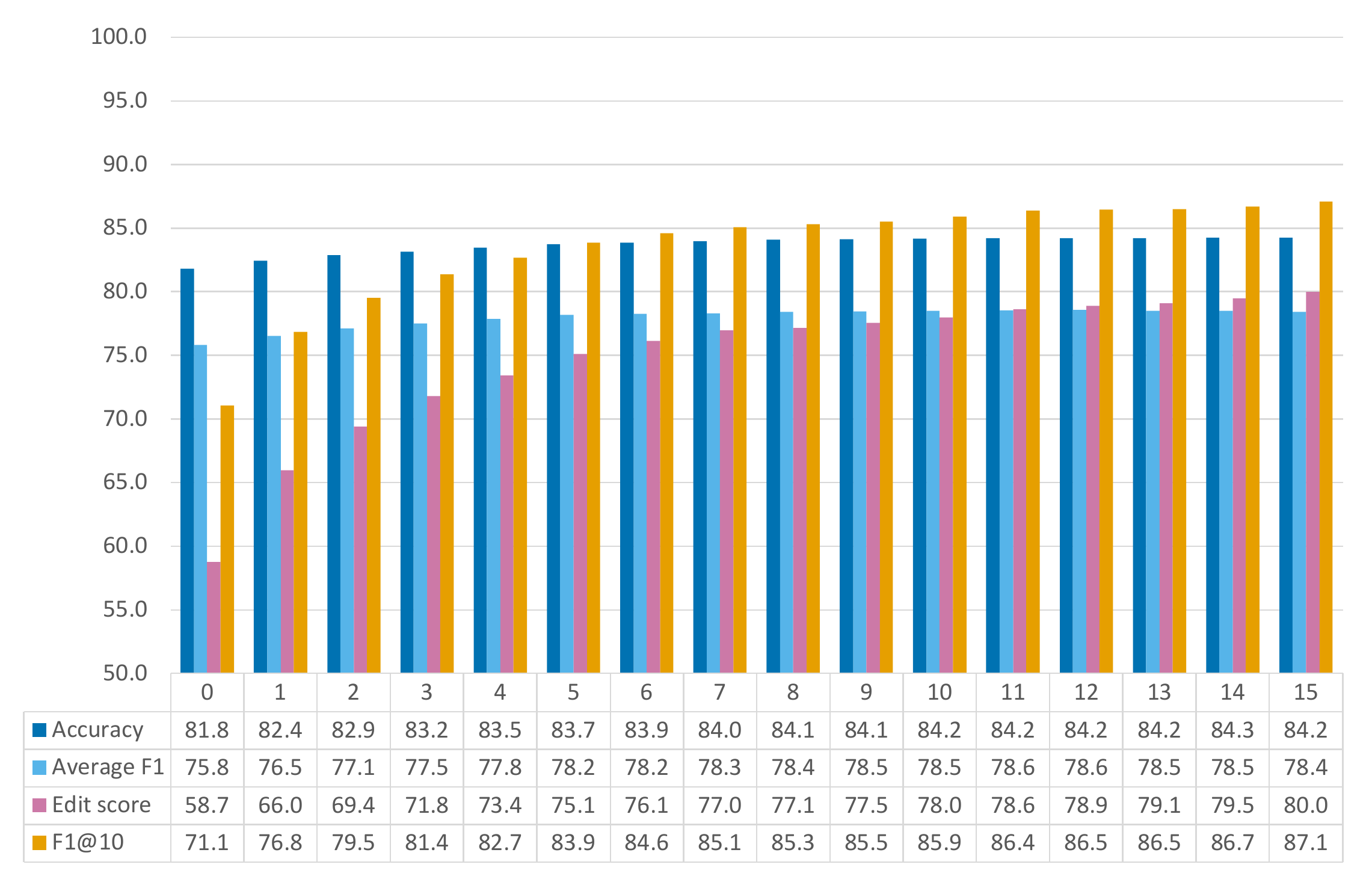}
\caption{Evaluating the sliding window approach with varying \emph{look ahead} times. For $k = 0, ..., 15$, we use $k$ future frames to estimate the gesture labels, i.e., $\bar{\hat{g}}(t) = \sum_{i = 0}^{k} \hat{g}_{t + i}(t)$. The numbers along the x-axis refer to different values of $k$.
The results were obtained at 5~fps using the 3D~CNN~(K) model. All measures are given in \%.} 
\end{figure}

\end{document}